\documentclass{article}
\pdfoutput=1

\usepackage{spconf}

\usepackage[utf8]{inputenc}

\usepackage{amsmath}
\usepackage{graphicx}
\usepackage{booktabs}
\usepackage{hyperref}
\usepackage{url}

\usepackage{paralist}
\usepackage[inline]{enumitem}
\usepackage{listings}
\usepackage{microtype}
\usepackage{multirow}

\usepackage[disable]{todonotes}

\usepackage{tikz}
\usetikzlibrary{positioning, shapes, fit, arrows}
\usetikzlibrary{external}

\title{Leveraging Weakly Supervised Data to Improve End-to-End Speech-to-Text Translation}
\name{\begin{tabular}{c}Ye Jia \quad Melvin Johnson \quad Wolfgang Macherey \quad Ron J. Weiss \quad Yuan Cao \\
Chung-Cheng Chiu \quad Naveen Ari \quad Stella Laurenzo \quad Yonghui Wu\end{tabular}}
\address{Google Research\\[4pt] {   \normalsize
  \texttt{\{jiaye,melvinp\}@google.com}
}}
\begin{document}
\ninept
\maketitle

\begin{abstract}
End-to-end Speech Translation (ST) models have many potential advantages when compared to the cascade of Automatic Speech Recognition (ASR) and text Machine Translation (MT) models, including lowered inference latency and the avoidance of error compounding.
However, the quality of end-to-end ST is often limited by a paucity of training data, since it is difficult to collect large parallel corpora of speech and translated transcript pairs.
Previous studies have proposed the use of pre-trained components and multi-task learning in order to benefit from weakly supervised training data, such as speech-to-transcript or text-to-foreign-text pairs.
In this paper, we demonstrate that using pre-trained MT or text-to-speech (TTS) synthesis models to convert weakly supervised data into speech-to-translation pairs for ST training can be more effective than multi-task learning.
Furthermore, we demonstrate that a high quality end-to-end ST model can be trained using \emph{only} weakly supervised datasets, and that synthetic data sourced from unlabeled monolingual text or speech can be used to improve performance.
Finally, we discuss methods for avoiding overfitting to synthetic speech with a quantitative ablation study.
\end{abstract}

\section{Introduction}
\label{sec:intro}

Recent advances in deep learning and more specifically in sequence-to-sequence modeling have led to dramatic improvements in ASR~\cite{chan2016listen, chiu2017state} \todo{more references?} and MT~\cite{sutskever2014sequence, cho2014properties, bahdanau2014neural, wu2016google, vaswani2017attention} tasks. These successes naturally led to attempts to construct end-to-end speech-to-text translation systems as a single neural network~\cite{berard2016listen, weiss2017sequence}. Such end-to-end systems have advantages over a traditional cascaded system that performs ASR and MT consecutively in that they
\begin{inparaenum}[1)]
\item naturally avoid compounding errors between the two systems;
\item can directly utilize prosodic cues from speech to improve translation;
\item have lower latency by avoiding inference with two models;
and
\item lower memory and computational resource usage.
\end{inparaenum}

However, training such an end-to-end ST model typically requires a large set of parallel speech-to-translation training data. Obtaining such a large dataset is significantly more expensive than acquiring data for ASR and MT tasks. This is often a limiting factor for the performance of such end-to-end systems. Recently explored techniques to mitigate this issue include multi-task learning \cite{weiss2017sequence, anastasopoulos2018tied} and pre-trained components \cite{berard2018end} in order to utilize weakly supervised data, i.e.\ speech-to-transcript or text-to-translation pairs, in contrast to fully supervised speech-to-translation pairs.

Although multi-task learning has been shown to bring significant quality improvements to end-to-end ST systems, it has two constraints which limit the performance of the trained ST model:
\begin{inparaenum}[1)]
\item the shared components have to compromise between multiple tasks, which can limit their performance on individual tasks;
\item for each training example,
the gradients are calculated for a single task,
parameters are therefore updated independently for each task,
which may lead to sub-optimal solution for the entire multi-task optimization problem.
\end{inparaenum}

In this paper, we train end-to-end ST models
on much larger datasets than previous work,
spanning up to 100 million training examples, including 1.3K hours of translated speech and 49K hours of transcribed speech. We confirm that multi-task learning and pre-training are still beneficial at such a large scale.
We demonstrate that performance of our end-to-end ST system can be significantly improved, even outperforming multi-task learning, by using a large amount of data synthesized from weakly supervised data such as typical ASR or MT training sets.
Similarly, we show that it is possible to train a high-quality end-to-end ST model without any fully supervised
training data by leveraging pre-trained components and data synthesized from weakly supervised datasets.
Finally, we demonstrate that data synthesized from fully unsupervised monolingual datasets can be used to improve end-to-end ST performance.

 \section{Related Work}
\label{sec:related}

Early work on speech translation typically used a cascade of an ASR model and an MT model \cite{ney1999speech, matusov2005integration, post2013improved}, giving the MT model access to the predicted probabilities and uncertainties from the ASR. Recent work has focused on training end-to-end ST in a single model \cite{berard2016listen, weiss2017sequence}.

In order to utilize both
fully supervised data
and also
weakly supervised data,
\cite{weiss2017sequence, anastasopoulos2018tied} use multi-task learning to train the ST model jointly with the ASR and/or the MT model. By doing so, both of them achieved better performance with the end-to-end model than the cascaded model.
\cite{berard2018end} conducts experiments on a larger 236 hour English-to-French dataset and pre-trains the encoder and decoder prior to multi-task learning, which further improves performance. However, the end-to-end model performs worse than the cascaded model in that work. \cite{bansal2018pre} shows that pre-training a speech encoder on one language can improve ST quality on a different source language.

Using TTS synthetic data for training speech translation
was a requirement when no direct parallel training data is available, such as in \cite{berard2016listen, kano2018structured}.
In contrast, we show that even when a large fully supervised training set is available, using synthetic training data from a high quality multi-speaker TTS system can further improve the performance of an end-to-end ST model.
Synthetic data has also been used to improve ASR performance. \cite{tjandra2018machine} builds a cycle chain between TTS and ASR models, in which the output from one model is used to help the training of the other. Instead of using TTS, \cite{renduchintala2018multi} synthesizes repeated phoneme sequence from unlabeled text to mimic temporal duration in acoustic input.  Similarly, \cite{Hayashi18} trains a pseudo-TTS model to synthesize the latent representation from a pre-trained ASR model from text, and uses it for data augmentation in ASR training.

The MT synthetic data in this work helps the system in a manner similar to knowledge distillation \cite{hinton2015distilling}, since the network is trained to predict outputs from a pretrained MT model. In contrast, synthesizing speech inputs using TTS is more similar to MT back-translation \cite{Sennrich16}.

 \section{Models}
\label{sec:model}

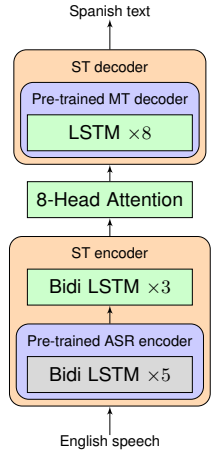
\begin{figure}[t]
  \centering

\scalebox{0.8}{
\begin{tikzpicture}[auto, font=\small\sffamily, node distance=2.2cm,auto,>=latex']

  \pgfdeclarelayer{back}
  \pgfdeclarelayer{middle}
  \pgfsetlayers{back,middle,main}

  \tikzstyle{cell} = [rectangle, draw, fill=green!20, text width=8em, text centered, minimum height=4.2ex]
  \tikzstyle{component} = [draw, minimum height=4.5ex, minimum width=6em, fill=blue!20, rounded corners=6]
  \tikzstyle{component_label} = [inner sep=2pt, align=left, font=\scriptsize\sffamily]

  \tikzstyle{io_label} = [align=center, inner sep=0]

  \node [io_label, name=input, align=center, font=\scriptsize\sffamily] {English speech};

  \node[cell, above=0.7cm of input, fill=gray!30] (asr_enc_cell) {Bidi LSTM $\times 5$};

  \node [component_label, above=0.09cm of asr_enc_cell] (asr_enc_label) {Pre-trained ASR encoder};
  \begin{pgfonlayer}{middle}
    \node[component, fit=(asr_enc_cell)(asr_enc_label)] (asr_enc) {};
  \end{pgfonlayer}

  \node[cell, above=0.3cm of asr_enc] (ast_enc_cell) {Bidi LSTM $\times 3$};

  \node [component_label, above=0.09cm of ast_enc_cell] (ast_enc_label) {ST encoder};
  \begin{pgfonlayer}{back}
    \node[component, fit=(asr_enc)(ast_enc_cell)(ast_enc_label), fill=orange!30] (ast_enc) {};
  \end{pgfonlayer}

  \node[cell, above=0.3cm of ast_enc] (attn) {8-Head Attention};

  \node[cell, above=0.5cm of attn] (MT_dec_cell) {LSTM $\times 8$};

  \node [component_label, above=0.09cm of MT_dec_cell] (MT_dec_label) {Pre-trained MT decoder};
  \begin{pgfonlayer}{middle}
    \node[component, fit=(MT_dec_cell)(MT_dec_label)] (MT_dec) {};
  \end{pgfonlayer}

  \node [component_label, above=0.09cm of MT_dec] (ast_dec_label) {ST decoder};
  \begin{pgfonlayer}{back}
    \node[component, fit=(MT_dec)(ast_dec_label), fill=orange!30] (ast_dec) {};
  \end{pgfonlayer}

  \node [io_label, above=0.5cm of ast_dec, font=\scriptsize\sffamily] (output) {Spanish text};

  \draw [->] (input) -- (ast_enc);
  \draw [->] (asr_enc) -- (ast_enc_cell);
  \draw [->] (ast_enc) -- (attn);
  \draw [->] (attn) -- (ast_dec);
  \draw [->] (ast_dec) -- (output);
\end{tikzpicture}
}

\caption{Overview of the end-to-end speech translation model. Blue blocks correspond to pre-trained components, grey components are frozen, and green components are fine-tuned on the ST task.}
\label{fig:model}

\end{figure}

Similar to \cite{weiss2017sequence}, we make use of three sequence-to-sequence models. Each one is composed of an encoder, a decoder, and an attention module. Besides the end-to-end ST model which is the major focus of this paper, we also build an ASR model and an MT model, which are used for building the baseline cascaded ST model, as well as for multi-task learning and encoder / decoder pre-training for ST.
All three models represent text using the same shared English/Spanish Word Piece Model (WPM) \cite{schuster2012japanese} containing 16K tokens.

\begin{itemize}[leftmargin=0em,label={}]
\item \textbf{ASR model:} Our ASR model follows the architecture of \cite{chiu2017state}. We use a 5 layer bidirectional LSTM encoder, with cell size 1024. The decoder is a 2 layer unidirectional LSTM with cell size 1024. The attention is 4-head additive attention. The model takes 80-channel log mel spectrogram features as input.
\item \textbf{MT model:} Our MT model follows the architecture of \cite{chen2018best}. We use a 6 layer bidirectional LSTM encoder, with cell size of 1024. The decoder is an 8 layer unidirectional LSTM with cell size 1024, with residual connection across layers. We use 8-head additive attention.
\item \textbf{ST model:} The encoder has similar architecture to the ASR encoder, and the decoder has similar architecture to the MT decoder. Throughout this work we experiment with varying the number of encoder layers. The model with the best performance is visualized in Figure \ref{fig:model}. It uses an 8 layer bidirectional LSTM for the encoder and an 8 layer unidirectional LSTM with residual connections for the decoder. The attention is 8-head additive attention, following the MT model.
\end{itemize}

\section{Synthetic training data}
\label{sec:synthetic}

Acquiring large-scale parallel speech-to-translation training data is extremely expensive.
The scarcity of such data is often a limiting factor on the quality of an end-to-end ST model.
To overcome this issue, we use two forms of weakly supervised data by:
synthesizing input speech corresponding to the input text in a parallel text MT training corpus, and synthesizing translated text targets from the output transcript in an ASR training corpus.

\subsection{Synthesis with TTS model}
\label{sec:synthetic-tts}

Recent TTS systems are able to synthesize speech with close to human naturalness \cite{shen2018natural}, in varied speakers' voices \cite{ping2017deep}, create novel voices by sampling from a continuous speaker embedding space \cite{jia2018transfer}.

In this work, we use the  TTS model trained on LibriSpeech~\cite{Libri15} from \cite{jia2018transfer}, except that we use a Griffin-Lim \cite{griffin1984signal} vocoder as in \cite{yx2017tacotron} which has significantly lower cost, but results in reduced audio quality\footnote{Synthetic waveforms are needed only for audio data augmentation. Otherwise, the mel-spectrogram predicted by the TTS model can be directly fed as input to the ST or ASR models, bypassing the vocoder.}. We  randomly sample from the continuous speaker embedding space for each synthesized 
example, resulting in wide diversity in speaker voices in the synthetic data.  This avoids unintentional bias toward a few synthetic speakers when using it to train an ST model and encourages generalization to speakers outside the training set.

\subsection{Synthesis with MT model}
\label{sec:synthetic-mt}

Another way to synthesize training data is to use an MT model to translate the transcripts in an ASR training set into the target language. In this work, we use the Google Translate service to obtain such translations. This procedure is similar to knowledge distillation \cite{hinton2015distilling}, except that it uses the final predictions as training targets rather than the predicted probability distributions. %
 \section{Experiments}
\label{sec:experiments}

\subsection{Datasets and metrics}

We focus on an English speech to Spanish text conversational speech translation task. Our experiments make use of three proprietary datasets, all consisting of conversational language, including: \begin{inparaenum}[(1)]
\item an MT training set of 70M English-Spanish parallel sentences;
\item an ASR training set of 29M hand-transcribed English utterances,
collected from anonymized voice search log;
and
\item a substantially smaller English-to-Spanish speech-to-translation set of 1M utterances obtained by sampling a subset from the 70M MT set, and crowd-sourcing humans to read the English sentences.
\end{inparaenum}
The final dataset can be directly used to train the end-to-end ST model.
We use data augmentation on both speech corpora by adding varying degrees of background noise and reverberation in the same manner as \cite{chiu2017state}.
The WPM shared among all models is trained with the 70M MT set.

We use two datasets for evaluation: a held out subset of 10.8K examples from the 1M ST set, which contains read speech, and another 8.9K recordings of natural conversational speech in a domain different from both the 70M MT and 29M ASR sets. Both eval sets contain English speech, English transcript and Spanish translation triples, so they can be used for evaluating either ASR, MT, or ST. ASR performance is measured in terms of Word Error Rate (WER) and translation performance is measured in terms of BLEU~\cite{papineni-EtAl:2002:ACL} scores, both on case and punctuation sensitive reference text.

\subsection{Baseline cascaded model}
\label{sec:cascaded}

We build a baseline system by training an English ASR model and an English-to-Spanish MT model with the architectures described in Section~\ref{sec:model}, and cascading them together by feeding the predicted transcript from the ASR model as the input to the MT model. The ASR model is trained on a mixture of the 29M ASR set and the 1M ST set with 8:1 per-dataset sampling probabilities in order to better adapt to the domain of the ST set. The MT model is trained on the 70M MT set, which is a superset of the 1M ST set.
As shown in Table~\ref{tbl:cascaded}, the ST BLEU is significantly lower than the MT BLEU, which is the result of cascading errors from the ASR model.

\begin{table}[t]
\centering
\begin{small}
\begin{tabular}{clcc}
    \toprule
    Task & Metric & In-domain & Out-of-domain \\
    \midrule
    ASR & WER\footnote{}    & 13.7\% & 30.7\% \\
    MT & BLEU & 78.8 & 35.6 \\
    ST & BLEU & 56.9 & 21.1 \\
    \bottomrule
\end{tabular}
\caption{Performance of the baseline cascaded ST system and the underlying ASR and MT components on both test sets.}
\label{tbl:cascaded}
\end{small}
\end{table}
\footnotetext{We report WER based on references which are case and punctuation sensitive in order to be consistent with the way BLEU is evaluated. The same ASR model obtains a WER of 6.9\% (in-domain) and 14.1\% (out-of-domain) if trained and evaluated on lower-cased transcripts without punctuation.}

\subsection{Baseline end-to-end models}
\label{sec:baselines}

We train a vanilla end-to-end ST model with a 5-layer encoder and an 8-layer decoder directly on the 1M ST set. We then adopt pre-training and multi-task learning as proposed in previous literature \cite{weiss2017sequence, anastasopoulos2018tied, berard2018end, bansal2018pre} in order to improve its performance. 
We pre-train the encoder on the ASR task, and the decoder on the MT task as described in 
Section~\ref{sec:cascaded}. After initialization with pre-trained components (or random values for components not being pre-trained), the ST model is fine-tuned on the 1M ST set.
Finally, we make use of multi-task learning by jointly training a combined network on the ST, ASR, and MT tasks, using the 1M, 29M, 70M datasets, respectively. The ST sub-network shares the encoder with the ASR network, and shares the decoder with the MT network. For each training step, one task is sampled and trained with equal probability.

\begin{table}[t]
\centering
\begin{small}
\begin{tabular}{lcc}
    \toprule
     & In-domain & Out-of-domain \\
    \midrule
    Cascaded & 56.9 & 21.1 \\
    \midrule
    Vanilla & 49.1 & 12.1 \\
    + Pre-training & 54.6 & 18.2 \\
    + Pre-training + Multi-task & 57.1 & 21.3 \\
    \bottomrule
\end{tabular}
\caption{BLEU scores of baseline end-to-end ST and cascaded models. All end-to-end models use 5 encoder and 8 decoder layers.}
\label{tbl:baselines}
\end{small}
\end{table}

Performance of these baseline models is shown in Table \ref{tbl:baselines}. 
Consistent with previous literature, we find that  pre-training and multi-task learning both significantly improve ST performance, because they increase the amount of data seen during training by two orders of magnitude.
When both pre-training and multi-task learning are applied, the end-to-end ST model slightly outperforms the cascaded model.

\subsection{Using synthetic training data}
\label{sec:exp-synthetic}

We explore the effect of using synthetic
training data as described in Section \ref{sec:synthetic}.
To avoid overfitting to TTS synthesized audio (especially since audio synthesized using the Griffin-Lim algorithm contains obvious and unnatural artifacts), we freeze the pre-trained encoder but stack a few additional layers on top of it. The impact of adding different number of additional layers when fine-tuning with the 1M ST set is shown in Table~\ref{tbl:additional-layers}.
Even with no additional layer, this approach outperforms the fully trainable model (Table \ref{tbl:baselines} row 3) on the out-of-domain eval set, which indicates that the frozen pre-trained encoder helps the ST model generalize better.
The benefit of adding extra
layers saturates after around 3 layers.
Following this result, we use 3 extra
layers for the following experiments, as visualized in Figure~\ref{fig:model}.

\begin{table}[t]
\centering
\begin{small}
\begin{tabular}{cccccc}
    \toprule
    & \multicolumn{5}{c}{\# additional encoder layers} \\
    & 0 & 1 & 2 & 3 & 4 \\
    \midrule
    In-domain     & 54.5 & 55.7 & 56.1 & 55.9 & 56.1  \\
    Out-of-domain & 19.5 & 18.8 & 19.3 & 19.5 & 19.6  \\
    \bottomrule
\end{tabular}
\caption{BLEU scores of the extended ST model, varying the number of additional encoder layers on top of the frozen pre-trained ASR encoder. The decoder is pre-trained but kept trainable. Multi-task learning is not used.}
\label{tbl:additional-layers}
\end{small}
\end{table}

\begin{table}[t]
\centering
\begin{small}
\begin{tabular}{lcc}
    \toprule
    Fine-tuning set & In-domain & Out-of-domain \\
    \midrule
    Real & 55.9 & 19.5 \\
    \midrule
    Real + TTS synthetic & 59.5 & 22.7 \\
    Real + MT synthetic  & 57.9 & 26.2 \\
    \textbf{Real + both synthetic} & \textbf{59.5} & \textbf{26.7} \\
    \midrule
    Only TTS synthetic & 53.9 & 20.8 \\
    Only MT synthetic  & 42.7 & 26.9 \\
    \textbf{Only both synthetic} & \textbf{55.6} & \textbf{27.0} \\
    \bottomrule
\end{tabular}
\caption{BLEU scores of ST trained with synthetic data. All rows use the same model architecture as Figure \ref{fig:model}.}
\label{tbl:synthetic}
\end{small}
\end{table}

We analyze the effect of using different synthetic datasets in Table~\ref{tbl:synthetic} for fine-tuning.
The TTS synthetic data is sourced from the 70M MT training set, by synthesizing English speech as described in Section~\ref{sec:synthetic-tts}. The synthesized speech is augmented with noise and reverberation following the same procedure as is used for real speech.
The MT synthetic data is sourced from the 29M ASR training set, by synthesizing translation to Spanish as described in Section \ref{sec:synthetic-mt}.
The encoder and decoder are both pre-trained on ASR and MT tasks, respectively, but multi-task learning is not used.

The middle group in Table \ref{tbl:synthetic} presents the result of fine-tuning with both synthetic datasets and the 1M real dataset, sampled with equal probability.
As expected, adding a large amount of synthetic training data (increasing the total number of training examples by 1 -- 2 orders of magnitude), significantly improves performance on both in-domain and out-of-domain eval sets.
The MT synthetic data  improves performance on the out-of-domain eval set more than it does on the in-domain set, partially because it contains natural speech instead of read speech, which is better matched to the out-of-domain eval set, and partially because it introduces more diversity to the training set and thus generalizes better.
Fine-tuning on the mixture of the three datasets results in dramatic gains on both eval sets, demonstrating that the two synthetic sources have complementary effects.
It also significantly outperforms the cascaded model.

The bottom group in Table~\ref{tbl:synthetic} fine-tunes using only synthetic data. Surprisingly, they achieve very good performance and even outperform training with both synthetic and real collected data on the out-of-domain eval set. This can be attributed to the increased sampling weight of training data
with natural speech (instead of read speech).
This result demonstrates the possibility of training a high-quality end-to-end ST system
with only weakly supervised data,
by using such data for components pre-training and generating synthetic parallel training data from them by leveraging on high quality TTS and MT models or services.

\subsection{Importance of frozen encoder and multi-speaker TTS}

To validate the importance of freezing the pre-trained encoder, we compare to a model where the encoder is fully fine-tuned on the ST task. As shown in Table~\ref{tbl:non-frozen}, full encoder fine-tuning hurts ST performance, by overfitting to the synthetic speech. The ASR encoder learns a high quality latent representation of the speech content when
pre-training on a large quantity of real speech with data augmentation.  Additional  fine-tuning on synthetic speech only hurts performance since the TTS data are not as realistic nor as diverse as real speech.

\begin{table}[t]
\centering
\begin{small}
\begin{tabular}{lcc}
    \toprule
    Fine-tuning set & In-domain & Out-of-domain \\
    \midrule
    Real + TTS synthetic & 58.7 & 21.4 \\
    Only TTS synthetic   & 35.1 &  9.8 \\
    \bottomrule
\end{tabular}
\caption{BLEU scores using fully trainable encoder, which performs worse than freezing lower encoder layers as in Table \ref{tbl:synthetic}.}
\label{tbl:non-frozen}
\end{small}
\end{table}

\begin{table}[t]
\centering
\begin{small}
\begin{tabular}{lcc}
    \toprule
    Fine-tuning set & \hspace{-1.5ex}In-domain\hspace{-1.5ex} & Out-of-domain \\
    \midrule
    Real + one-speaker TTS synthetic & 59.5 & 19.5 \\
    Only one-speaker TTS synthetic & 38.5 & 13.8 \\
    \bottomrule
\end{tabular}
\caption{BLEU scores when fine-tuning with synthetic speech data synthesized using a single-speaker TTS system, which performs worse than using the multi-speaker TTS as in Table \ref{tbl:synthetic}.}
\label{tbl:single-speaker}
\end{small}
\end{table}

Similarly, to validate the importance of using a high quality multi-speaker TTS system to synthesize training data with wide speaker variation, we train models using data synthesized with the single speaker TTS model from \cite{shen2018natural}.
This model generates more natural speech than
the multi-speaker model used in Sec.~\ref{sec:exp-synthetic} \cite{jia2018transfer}.
To ensure a fair comparison, we use a Griffin-Lim vocoder and
 a 16 kHz sampling rate.
We use the same data augmentation procedure described above.

Results are shown in Table~\ref{tbl:single-speaker}. Even though a frozen pre-trained encoder is used, fine-tuning on only single-speaker TTS synthetic data still performs much worse than fine-tuning with multi-speaker TTS data, especially on the out-of-domain eval set.
However, when trained on the combination of real and synthetic speech, performance 
on the in-domain eval set is not affected.
We conjecture that this is because the in-domain eval set consists of read speech, which is better matched to the prosodic quality of the single-speaker TTS model.
The large performance degradation on the out-of-domain eval set again indicates worse generalization.

Incorporating recent advances in TTS to introduce more natural prosody and style variation \cite{skerry2018towards,wang2018style,hsu2018hierarchical} to the synthetic speech might further improve performance when training on synthetic speech. We leave such investigations as future work.

\subsection{Utilizing unlabelled monolingual text or speech}

In this section, we go further and show that unlabeled monolingual speech and text can be leveraged to improve performance of an end-to-end ST model, by using them to synthesize parallel speech-to-translation examples using available ASR, MT, and TTS systems. Even though such datasets are highly synthetic, they can still  benefit an ST model trained with as many as 1M real training examples.

We take the English text from the 70M MT set as an unlabeled text set, synthesize English speech for it using a multi-speaker TTS model as in Section \ref{sec:exp-synthetic}, and translate it to Spanish using the Google Translate service. 
Similarly, we take the English speech from the 29M ASR set as an unlabeled speech set, synthesize translation targets for it by using the cascaded model we build in Section \ref{sec:cascaded}. We use this cascaded model only to enable comparison to its own performance. Replacing it with a cascade of other ASR and MT models or services should not change the conclusion.
Since there is no parallel training data for ASR or MT in this case, pre-training does not apply. We use the vanilla model with a 5-layer encoder as in Section \ref{sec:baselines}.

\begin{table}[t]
\centering
\begin{small}
\begin{tabular}{lcc}
    \toprule
    Training set & In-domain & Out-of-domain \\
    \midrule
    Real & 49.1 & 12.1 \\
    Real + Synthetic from text   & 55.9 & 19.4 \\
    Real + Synthetic from speech & 52.4 & 15.3 \\
    Real + Synthetic from both   & 55.8 & 16.9 \\
    \bottomrule
\end{tabular}
\caption{BLEU scores for the vanilla model trained on synthetic data generated from unlabeled monolingual data, without pre-training.}
\label{tbl:unlabeled}
\end{small}
\end{table}

Results are presented in Table~\ref{tbl:unlabeled}.
Even though these datasets are highly synthetic, they still significantly improve performance over the vanilla model.
Because the unlabeled speech is processed with weaker models, it doesn't bring as much gain as the synthetic set from unlabeled text. Since it essentially distills knowledge from the cascaded model, it is also understandable that it does not outperform it.
Performance is far behind our best results in Section~\ref{sec:exp-synthetic} since pre-training is not used.
Nevertheless, this result demonstrates that with access to high-quality ASR, MT, and/or TTS systems, one can leverage large sets of unlabeled monolingual data to improve the quality of an end-to-end ST system, even if a small amount of direct parallel training data are available. \section{Conclusions}
\label{sec:conslusions}

We propose a weakly supervised learning procedure that leverages synthetic training data to fine-tune an end-to-end sequence-to-sequence
ST
model, whose encoder and decoder networks have been separately pre-trained on
ASR and MT
tasks, respectively.
We demonstrate that this approach outperforms multi-task learning in experiments on a large scale English speech to Spanish text translation task.
When utilizing synthetic speech inputs, we find that it is important to use a high quality multispeaker TTS model, and to freeze the pre-trained encoder to avoid overfitting to synthetic audio.
We explore even more impoverished data scenarios, and 
show that it is possible to train a high quality end-to-end ST model by fine-tuning \emph{only} on synthetic data from readily available ASR or MT training sets. 
Finally, we demonstrate that a large quantity of unlabeled speech or text can be leveraged to improve an end-to-end ST model when a small fully supervised
training corpus is available.

\section{Acknowledgments}

The authors thank Patrick Nguyen, Orhan Firat, the Google Brain team, the Google Translate team, and the Google Speech Research team for their helpful discussions and feedback, as well as Mengmeng Niu for her operational support on data collection.

\bibliographystyle{IEEEbib}
\bibliography{references}

\begin{thebibliography}{10}

\bibitem{chan2016listen}
W.~Chan, N.~Jaitly, Q.~Le, and O.~Vinyals,
\newblock ``Listen, attend and spell: A neural network for large vocabulary
  conversational speech recognition,''
\newblock in {\em Proc. ICASSP}, 2016.

\bibitem{chiu2017state}
C.-C. Chiu, T.~N. Sainath, Y.~Wu, R.~Prabhavalkar, P.~Nguyen, Z.~Chen,
  A.~Kannan, R.~J. Weiss, K.~Rao, K.~Gonina, et~al.,
\newblock ``State-of-the-art speech recognition with sequence-to-sequence
  models,''
\newblock in {\em Proc. ICASSP}, 2018.

\bibitem{sutskever2014sequence}
I.~Sutskever, O.~Vinyals, and Q.~V. Le,
\newblock ``Sequence to sequence learning with neural networks,''
\newblock in {\em Advances in NeurIPS}, 2014.

\bibitem{cho2014properties}
K.~Cho, B.~Van~Merri{\"e}nboer, D.~Bahdanau, and Y.~Bengio,
\newblock ``On the properties of neural machine translation: Encoder-decoder
  approaches,''
\newblock in {\em Eighth Workshop on Syntax, Semantics and Structure in
  Statistical Translation}, 2014.

\bibitem{bahdanau2014neural}
D.~Bahdanau, K.~Cho, and Y.~Bengio,
\newblock ``Neural machine translation by jointly learning to align and
  translate,''
\newblock in {\em Proc. ICLR}, 2015.

\bibitem{wu2016google}
Y.~Wu, M.~Schuster, Z.~Chen, Q.~V. Le, M.~Norouzi, W.~Macherey, M.~Krikun,
  Y.~Cao, Q.~Gao, K.~Macherey, et~al.,
\newblock ``Google's neural machine translation system: Bridging the gap
  between human and machine translation,''
\newblock {\em arXiv preprint arXiv:1609.08144}, 2016.

\bibitem{vaswani2017attention}
A.~Vaswani, N.~Shazeer, N.~Parmar, J.~Uszkoreit, L.~Jones, A.~N. Gomez,
  {\L}.~Kaiser, and I.~Polosukhin,
\newblock ``Attention is all you need,''
\newblock in {\em Advances in NeurIPS}, 2017.

\bibitem{berard2016listen}
A.~B{\'e}rard, O.~Pietquin, C.~Servan, and L.~Besacier,
\newblock ``Listen and translate: A proof of concept for end-to-end
  speech-to-text translation,''
\newblock in {\em NeurIPS Workshop on End-to-end Learning for Speech and Audio
  Processing}, 2016.

\bibitem{weiss2017sequence}
R.~J. Weiss, J.~Chorowski, N.~Jaitly, Y.~Wu, and Z.~Chen,
\newblock ``Sequence-to-sequence models can directly translate foreign
  speech,''
\newblock in {\em Proc. Interspeech}, 2017.

\bibitem{anastasopoulos2018tied}
A.~Anastasopoulos and D.~Chiang,
\newblock ``Tied multitask learning for neural speech translation,''
\newblock in {\em Proc. NAACL-HLT}, 2018.

\bibitem{berard2018end}
A.~B{\'e}rard, L.~Besacier, A.~C. Kocabiyikoglu, and O.~Pietquin,
\newblock ``End-to-end automatic speech translation of audiobooks,''
\newblock in {\em Proc. ICASSP}, 2018.

\bibitem{ney1999speech}
H.~Ney,
\newblock ``Speech translation: Coupling of recognition and translation,''
\newblock in {\em Proc. ICASSP}, 1999.

\bibitem{matusov2005integration}
E.~Matusov, S.~Kanthak, and H.~Ney,
\newblock ``On the integration of speech recognition and statistical machine
  translation,''
\newblock in {\em European Conference on Speech Communication and Technology},
  2005.

\bibitem{post2013improved}
M.~Post, G.~Kumar, A.~Lopez, D.~Karakos, C.~Callison-Burch, and S.~Khudanpur,
\newblock ``Improved speech-to-text translation with the {Fisher and Callhome
  Spanish--English} speech translation corpus,''
\newblock in {\em Proc. IWSLT}, 2013.

\bibitem{bansal2018pre}
S.~Bansal, H.~Kamper, K.~Livescu, A.~Lopez, and S.~Goldwater,
\newblock ``Pre-training on high-resource speech recognition improves
  low-resource speech-to-text translation,''
\newblock {\em arXiv preprint arXiv:1809.01431}, 2018.

\bibitem{kano2018structured}
T.~Kano, S.~Sakti, and S.~Nakamura,
\newblock ``Structured-based curriculum learning for end-to-end
  english-japanese speech translation,''
\newblock in {\em Proc. Interspeech}, 2017.

\bibitem{tjandra2018machine}
A.~Tjandra, S.~Sakti, and S.~Nakamura,
\newblock ``Machine speech chain with one-shot speaker adaptation,''
\newblock in {\em Proc. Interspeech}, 2018.

\bibitem{renduchintala2018multi}
A.~Renduchintala, S.~Ding, M.~Wiesner, and S.~Watanabe,
\newblock ``Multi-modal data augmentation for end-to-end {ASR},''
\newblock in {\em Proc. Interspeech}, 2018.

\bibitem{Hayashi18}
T.~Hayashi, S.~Watanabe, Y.~Zhang, T.~Toda, T.~Hori, R.~Astudillo, and
  K.~Takeda,
\newblock ``Back-translation-style data augmentation for end-to-end {ASR},''
\newblock {\em arXiv preprint arXiv:1807.10893}, 2018.

\bibitem{hinton2015distilling}
G.~Hinton, O.~Vinyals, and J.~Dean,
\newblock ``Distilling the knowledge in a neural network,''
\newblock in {\em NeurIPS Deep Learning and Representation Learning Workshop},
  2015.

\bibitem{Sennrich16}
R.~Sennrich, B.~Haddow, and A.~Birch,
\newblock ``Improving neural machine translation models with monolingual
  data,''
\newblock in {\em Proc. Association for Computational Linguistics (ACL)}, 2016.

\bibitem{schuster2012japanese}
M.~Schuster and K.~Nakajima,
\newblock ``Japanese and {Korean} voice search,''
\newblock in {\em Proc. ICASSP}, 2012.

\bibitem{chen2018best}
M.~X. Chen, O.~Firat, A.~Bapna, M.~Johnson, W.~Macherey, G.~Foster, L.~Jones,
  N.~Parmar, M.~Schuster, Z.~Chen, Y.~Wu, and M.~Hughes,
\newblock ``The best of both worlds: Combining recent advances in neural
  machine translation,''
\newblock in {\em Proc. Association for Computational Linguistics (ACL)}, 2018.

\bibitem{shen2018natural}
J.~Shen, R.~Pang, R.~J. Weiss, M.~Schuster, N.~Jaitly, Z.~Yang, Z.~Chen,
  Y.~Zhang, Y.~Wang, R.~Skerry-Ryan, et~al.,
\newblock ``Natural {TTS} synthesis by conditioning wavenet on mel spectrogram
  predictions,''
\newblock in {\em Proc. ICASSP}, 2017.

\bibitem{ping2017deep}
W.~Ping, K.~Peng, A.~Gibiansky, S.~O. Arik, A.~Kannan, S.~Narang, J.~Raiman,
  and J.~Miller,
\newblock ``Deep voice 3: 2000-speaker neural text-to-speech,''
\newblock in {\em Proc. ICLR}, 2018.

\bibitem{jia2018transfer}
Y.~Jia, Y.~Zhang, R.~J. Weiss, Q.~Wang, J.~Shen, F.~Ren, Z.~Chen, P.~Nguyen,
  R.~Pang, I.~L. Moreno, and Y.~Wu,
\newblock ``Transfer learning from speaker verification to multispeaker
  text-to-speech synthesis,''
\newblock in {\em Advances in NeurIPS}, 2018.

\bibitem{Libri15}
V.~Panayotov, G.~Chen, D.~Povey, and S.~Khudanpur,
\newblock ``{LibriSpeech}: an {ASR} corpus based on public domain audio
  books,''
\newblock in {\em Proc. ICASSP}, 2015.

\bibitem{griffin1984signal}
D.~Griffin and J.~Lim,
\newblock ``Signal estimation from modified short-time {F}ourier transform,''
\newblock {\em IEEE Transactions on Acoustics, Speech, and Signal Processing},
  vol. 32, no. 2, 1984.

\bibitem{yx2017tacotron}
Y.~Wang, R.~Skerry-Ryan, D.~Stanton, Y.~Wu, R.~J. Weiss, N.~Jaitly, Z.~Yang,
  Y.~Xiao, Z.~Chen, S.~Bengio, Q.~Le, Y.~Agiomyrgiannakis, R.~Clark, et~al.,
\newblock ``Tacotron: Towards end-to-end speech synthesis,''
\newblock in {\em Proc. Interspeech}, 2017.

\bibitem{papineni-EtAl:2002:ACL}
K.~Papineni, S.~Roukos, T.~Ward, and W.-J. Zhu,
\newblock ``{BLEU}: A method for automatic evaluation of machine translation,''
\newblock in {\em Proc. Association for Computational Linguistics (ACL)}, 2002.

\bibitem{skerry2018towards}
R.~Skerry-Ryan, E.~Battenberg, Y.~Xiao, Y.~Wang, D.~Stanton, J.~Shor, R.~J.
  Weiss, R.~Clark, and R.~A. Saurous,
\newblock ``Towards end-to-end prosody transfer for expressive speech synthesis
  with {T}acotron,''
\newblock in {\em Proc. ICML}, 2018.

\bibitem{wang2018style}
Y.~Wang, D.~Stanton, Y.~Zhang, R.~Skerry-Ryan, E.~Battenberg, J.~Shor, Y.~Xiao,
  F.~Ren, Y.~Jia, and R.~A. Saurous,
\newblock ``Style tokens: Unsupervised style modeling, control and transfer in
  end-to-end speech synthesis,''
\newblock in {\em Proc. ICML}, 2018.

\bibitem{hsu2018hierarchical}
W.-N. Hsu, Y.~Zhang, R.~J. Weiss, H.~Zen, Y.~Wu, Y.~Wang, Y.~Cao, Y.~Jia,
  Z.~Chen, J.~Shen, P.~Nguyen, and R.~Pang,
\newblock ``Hierarchical generative modeling for controllable speech
  synthesis,''
\newblock in {\em Proc. ICLR}, 2019,
\newblock to appear.

\end{thebibliography}

\end{document}